%% file: main.tex
\documentclass[10pt,twocolumn,letterpaper]{article}

\usepackage{cvpr}              % To produce the CAMERA-READY version

\usepackage{algorithm}
\usepackage{algpseudocode}
\DeclareMathOperator*{\E}{\mathbb{E}} % Expectation
\DeclareMathOperator*{\KL}{KL}       % KL Divergence

\newcommand{\loss}{\mathcal{L}}
\newcommand{\data}{\mathcal{D}}
\newcommand{\params}{\theta}
\newcommand{\weights}{W}
\newcommand{\pathvec}{B}

\input{preamble}
\definecolor{cvprblue}{rgb}{0.21,0.49,0.74}
\usepackage[pagebackref,breaklinks,colorlinks,allcolors=cvprblue]{hyperref}

\def\paperID{}  %45246} % *** Enter the Paper ID here
\def\confName{CVPR}
\def\confYear{2026}

\title{Monte Carlo Stochastic Depth for Uncertainty Estimation in Deep Learning}

\author{Adam T. Müller \quad Tobias Rögelein \quad Nicolaj C. Stache\\
Heilbronn University of Applied Sciences, Germany\\
{\tt\small \{adam-theo.mueller, tobias.roegelein, nicolaj.stache\}@hs-heilbronn.de}
}

\begin{document}
\maketitle
\input{sec/0_abstract}

\input{sec/1_intro}

\input{sec/2_relWork}

\input{sec/3_background}

\input{sec/4_MCSD}

\input{sec/5_experiments}

\input{sec/6_limitations}

\input{sec/7_conclusion}

%\newpage
{
    \small
    \bibliographystyle{ieeenat_fullname}
    \bibliography{main}
}

\input{sec/A_sup_Reproducibility}

\input{sec/B_supp_theory}

\input{sec/CD_sup_AddResultsDiscussion}

\end{document}

%% file: sec/0_abstract.tex
\begin{abstract}
The deployment of deep neural networks in safety-critical systems necessitates reliable and efficient uncertainty quantification (UQ). A practical and widespread strategy for UQ is repurposing stochastic regularizers as scalable approximate Bayesian inference methods, such as Monte Carlo Dropout (MCD) and MC-DropBlock (MCDB). However, this paradigm remains under-explored for Stochastic Depth (SD), a regularizer integral to the residual-based backbones of most modern architectures. While prior work demonstrated its empirical promise for segmentation, a formal theoretical connection to Bayesian variational inference and a benchmark on complex, multi-task problems like object detection are missing.

\noindent In this paper, we first provide theoretical insights connecting Monte Carlo Stochastic Depth (MCSD) to principled approximate variational inference. We then present the first comprehensive empirical benchmark of MCSD against MCD and MCDB on state-of-the-art detectors (YOLO, RT-DETR) using the COCO and COCO-O datasets. Our results position MCSD as a robust and computationally efficient method that achieves highly competitive predictive accuracy (mAP), notably yielding slight improvements in calibration (ECE) and uncertainty ranking (AUARC) compared to MCD. We thus establish MCSD as a theoretically-grounded and empirically-validated tool for efficient Bayesian approximation in modern deep learning.

\end{abstract}

%% file: sec/1_intro.tex
\section{Introduction}
\label{sec:intro}

The deployment of deep neural networks (DNNs) in safety-critical applications such as autonomous driving, robotics, and medical diagnostics poses a present-day engineering challenge. While these models have achieved remarkable predictive accuracy~\cite{gui_fiptr_2025, de_fauw_clinically_2018}, their standard deterministic nature is a fundamental limitation~\cite{guo_calibration_2017, patrone_probabilistic_2025}. Such models are often poorly calibrated and can produce high-confidence predictions for inputs far from their training data, failing to signal uncertainty for out-of-distribution (OOD) inputs~\cite{guo_calibration_2017}. This propensity for high-confidence errors is unacceptable in high-stakes environments where an overconfident error can be catastrophic. Consequently, equipping such models with reliable, computationally-efficient uncertainty quantification is a critical and necessary step for their trustworthy deployment.

The Bayesian framework~\cite{wang_survey_2020, papamarkou_position_2024} offers a principled solution by learning a posterior distribution over model parameters, naturally capturing epistemic uncertainty. However, exact Bayesian inference in deep networks is computationally intractable. A practical breakthrough was achieved by Gal, Y. and Ghahramani, Z.~\cite{gal_dropout_2016}, who re-framed stochastic regularization as a form of approximate Bayesian variational inference. Their method, Monte Carlo Dropout (MCD), applies dropout at inference time to sample from the model's approximate posterior, providing a scalable way to estimate epistemic uncertainty. This paradigm of repurposing a regularizer for Bayesian approximation was later extended~\cite{gal_concrete_2017, mukhoti_evaluating_2019, zeevi_ratein_2025}. As standard dropout is less effective in convolutional networks, Monte Carlo DropBlock (MCDB) was proposed by Yelleni, S. H. \etal~\cite{yelleni_monte_2024}, applying the same Monte Carlo (MC) principle to the more suitable, architecture-aware DropBlock regularizer.

This progression from MCD to MCDB reveals a powerful meta-strategy: stochastic regularization techniques implicitly define an approximate posterior, which can be sampled at inference time. This naturally leads to a new research question: can this principle be extended to other, more architecturally-ingrained regularizers? Stochastic Depth (SD)~\cite{huang_deep_2016} is a prime candidate. It is a powerful regularization technique applicable to the residual-based architectures that form the backbones of most modern, high-performance models, including state-of-the-art object detectors (e.g. YOLO)~\cite{Terven2023YOLOs} and Vision Transformer (ViT) architectures~\cite{han_survey_2023}. Repurposing SD for inference-time sampling would create an implicit ensemble of models with varying depths, a stochastic process that is fundamentally different from the unit-dropping of MCD or the region-dropping of MCDB.

This idea has recently seen preliminary validation. Yao, L. \etal~\cite{Yao_SDinVI_2024} provided the first empirical evidence that Bayesian inference based on SD can yield well-calibrated uncertainty estimates, demonstrating its promise for semantic segmentation. However, this prior work leaves two critical gaps, which our paper addresses. First, the work remains purely empirical, lacking the formal theoretical connection to Bayesian variational inference -- a limitation acknowledged in the original study~\cite{Yao_SDinVI_2024}. Second, its performance on complex, multi-task problems like object detection remains unexplored.

This paper bridges these gaps. We provide theoretical insights positioning Monte Carlo Stochastic Depth (MCSD) as a valid approximate Bayesian inference technique. We then conduct the first comprehensive empirical benchmark of MCSD for object detection, validating it against MCD and MCDB on state-of-the-art architectures. Our findings establish MCSD as a theoretically-grounded and empirically-validated tool for efficient Bayesian approximation in modern deep learning.

Our main contributions are as follows: \textbf{(1)} We formally derive that MCSD is equivalent to a variational approximation of a Bayesian neural network. \textbf{(2)} We adapt and validate MCSD for the task of object detection, conducting its first empirical study in this domain. \textbf{(3)} We provide comprehensive experiments of MCSD against MCD and MCDB on current architectures (FasterRCNN~\cite{ren_faster_2015}, YOLO~\cite{redmon_you_2016}, RT-DETR~\cite{zhao_detrs_2024}), establishing compatibility with multiple Deep Neural Network (DNN) architectures, provided they are equipped with skip-connections.

%% file: sec/2_relWork.tex
\section{Related Work}
\label{sec:relWork}

Deploying deep neural networks in high-stakes environments necessitates rigorous predictive uncertainty quantification~\cite{begoli_need_2019}. This uncertainty is conventionally decomposed into irreducible data noise (\textbf{aleatoric}) and reducible model ignorance due to a lack of data (\textbf{epistemic})~\cite{kendall_what_2017, hora_aleatory_1996, hullermeier_aleatoric_2021, kiureghian_aleatory_2009}. Our work focuses on estimating epistemic uncertainty. This remains a central challenge, as it requires characterizing the high-dimensional posterior weight distribution, $p(W \rvert \mathcal{D})$, a task that is analytically and computationally intractable for modern DNNs~\cite{blundell_weight_2015}.

\vspace{0.13cm}
\noindent\textbf{Bayesian Approximation and Ensemble Methods.} Bayesian Neural Networks (BNNs)~\cite{goan_bayesian_2020, blundell_weight_2015} provide a principled framework for capturing this uncertainty by learning a distribution over weights rather than point estimates. As the exact posterior is intractable, scalable approximation methods are required. These can be broadly divided into:
\vspace{0.13cm}
\begin{itemize}
  \item \textbf{Markov Chain Monte Carlo (MCMC):}~\cite{ma_complete_2015, welling_bayesian_2011, neal_mcmc_2011} These methods can, in theory, converge to the true posterior but are computationally prohibitive and too slow for deep learning.
  \item \textbf{Variational Inference (VI):} This family of methods~\cite{ganguly_introduction_2021, blei_variational_2017} approximates the true posterior $p(W \rvert \mathcal{D})$ with a simpler, tractable variational distribution $q_{\theta}(W)$ (e.g., a diagonal Gaussian). This optimization-based approach is far more scalable and, as we will detail in~\cref{sec:background}, forms the theoretical basis for many modern techniques.
\end{itemize}
\vspace{0.13cm}

\noindent While BNNs offer a formal approach, Deep Ensembles~\cite{lakshminarayanan_simple_2017, rahaman_uncertainty_2021} have emerged as an empirically powerful and robust method for uncertainty quantification. This non-Bayesian method trains multiple identical models with different random initializations and treats the variance of their predictions as uncertainty. However, Ensemble methods incur a significant computational burden, requiring N-times the cost for both training and inference.

\vspace{0.13cm}
\noindent\textbf{Post-Hoc and Monte Carlo Approximations.} As a computationally efficient alternative to ensembles, post-hoc methods have garnered significant interest~\cite{slack_reliable_2021}. These approaches aim to extract uncertainty from a single, pre-trained model without costly re-training. This is especially valuable for large-scale computer vision applications, where model and dataset sizes continue to grow, as it provides a way to equip large models with uncertainty quantification (UQ)~\cite{franchi_make_2023}.
\vspace{0.13cm}
\begin{itemize}
  \item \textbf{Laplace Approximation}~\cite{zhdanov_identity_2025, bergamin_riemannian_2023} is a post-hoc technique that fits a Gaussian to the posterior centered at a Maximum A Posteriori (MAP) estimate, using the local curvature (Hessian) of the loss. While scalable variants like last-layer or diagonal approximations exist~\cite{yun_laplace_2023, ritter_scalable_2018}, their utility can be limited.
  \item \textbf{Stochastic Regularization for Bayesian Approximation} is a paradigm more central to our work. Gal, Y. and Ghahramani, Z.~\cite{gal_dropout_2016} provided the critical insight that a standard NN trained with dropout (or generalized weight-dropping variants like DropConnect~\cite{wan_regularization_2013}) is equivalent to an approximation of a deep Gaussian Process. Consequently, performing multiple stochastic forward passes at inference time constitutes a form of variational inference. This principle has been extended to other stochastic techniques, such as MC-DropBlock~\cite{yelleni_monte_2024} and Stochastic Batch Normalization~\cite{atanov_uncertainty_2019}. Seminal to our work, Yao, L. \etal~\cite{Yao_SDinVI_2024} introduced the idea of VI with SD as a Bayesian approximation. Concurrently, Depth Uncertainty Networks (DUNs)~\cite{antoran_depth_2020} have explored treating network depth as a categorical random variable, marginalizing predictions over subnetworks of increasing depth in a single forward pass.
\end{itemize}
\vspace{0.13cm}

\noindent\textbf{Single-Pass Uncertainty.} Distinct from sampling techniques, deterministic methods quantify uncertainty in a single forward pass. Approaches like Evidential Deep Learning~\cite{sensoy_evidential_2018} and Epistemic Neural Networks~\cite{osband_epistemic_2023} are highly efficient but require specialized loss functions or architectural modifications. Conversely, MCSD leverages native stochastic regularizers, requiring no structural overhead or custom training regimes.

\vspace{0.13cm}
\noindent\textbf{Uncertainty in Object Detection.} Beyond standard classification, UQ is increasingly vital for object detection. Prior works adapt MCD to extract spatial and label uncertainty for improved open-set performance~\cite{miller_dropout_2018} and corner-case detection~\cite{heidecker_towards_2021}. Our evaluation of MCSD advances this domain by offering an efficient, architecturally native alternative to MCD.

\vspace{0.13cm}
\noindent Building upon these insights, we present the theoretical properties of Bayesian VI using stochastic depth. We propose MCSD as a novel stochastic process, that yields robust uncertainty estimates in deployment across a variety of model architectures. To provide mathematical grounding for both established methods like MCD and the proposed MCSD, the following section presents the theoretical background on VI and MC sampling.

%% file: sec/3_background.tex
\section{Background}
\label{sec:background}

In a standard BNN, the aim is to learn a full posterior distribution $p(W \rvert \mathcal{D})$ over the weights given the training data $\mathcal{D}$. Making a prediction $y_*$ for a new input $x_*$ involves marginalizing over this posterior, an operation known as Bayesian model averaging~\cite{seoh_qualitative_2020}:
\begin{equation}
 p(y_{*} \rvert x_{*}, \mathcal{D}) = \int p(y_{*} \rvert x_{*}, W) p(W \rvert \mathcal{D}) dW
 \label{eq:pred_posterior}
\end{equation}
Here, $p(y_{*} \rvert x_{*}, W)$ is the likelihood of the prediction given a specific set of weights, and $p(W \rvert \mathcal{D})$ is the posterior distribution of the weights given the data. However, as noted in \cref{sec:relWork}, the posterior $p(W \rvert \mathcal{D})$ is intractable for any non-trivial deep neural network (DNN).

\subsection{Variational Inference}
VI reformulates this intractable integration problem as an optimization problem~\cite{ganguly_introduction_2021}. The core idea is to introduce a simpler, tractable variational distribution $q_{\theta}(W)$, to approximate the true posterior. The goal is to find the parameters $\theta$ that minimize the distance between $q_{\theta}(W)$ and $p(W \rvert \mathcal{D})$.

This is conventionally achieved by minimizing the \textbf{Kullback-Leibler (KL) divergence} between the approximation and the true posterior~\cite{acerbi_variational_2018}:
\begin{equation}
 \theta^{*} = \arg \min_{\theta} \text{KL}(q_{\theta}(W) || p(W \rvert \mathcal{D}))
 \label{eq:kl_minimization}
\end{equation}
Minimizing this KL divergence is still intractable because it depends on the posterior. However, it can be shown to be equivalent to maximizing a different, tractable objective: the \textbf{Evidence Lower Bound (ELBO)}, denoted $\mathcal{L}_{\text{VI}}(\theta)$~\cite{acerbi_variational_2018}:
\begin{equation}
 \log p(\mathcal{D}) = \text{KL}(q_{\theta}(W) || p(W \rvert \mathcal{D})) + \mathcal{L}_{\text{VI}}(\theta)
 \label{eq:elbo_identity}
\end{equation}
Since the evidence $\log p(\mathcal{D})$ is constant with respect to $\theta$, maximizing the ELBO is equivalent to minimizing the KL divergence. The ELBO itself is defined as~\cite{ganguly_introduction_2021}:
\begin{equation}
 \mathcal{L}_{\text{VI}}(\theta) = \underbrace{\mathbb{E}_{q_{\theta}(W)}[\log p(\mathcal{D} \rvert W)]}_{\text{Expected Log-Likelihood}} - \underbrace{\text{KL}(q_{\theta}(W) || p(W))}_{\text{Complexity Penalty}}
 \label{eq:elbo_objective}
\end{equation}
Here, $p(W)$ is the prior distribution over the weights, making this objective tractable. The first term encourages the distribution $q_{\theta}(W)$ to find weights that explain the data well, while the second term (the KL penalty) acts as a regularizer, forcing the approximation to stay close to the prior.

\subsection{Bayesian Prediction via Monte Carlo Sampling}
Once $q_{\theta}(W)$ is optimized by maximizing the ELBO, it can be used to approximate the predictive posterior from \cref{eq:pred_posterior}:
\begin{equation}
 p(y_{*} \rvert x_{*}, \mathcal{D}) \approx \int p(y_{*} \rvert x_{*}, W) q_{\theta}(W) dW
 \label{eq:approx_pred_posterior}
\end{equation}
While $q_{\theta}(W)$ is tractable, this integral often is not. It is therefore approximated using MC sampling~\cite{franchi_make_2023}. $T$ sets of weights $\{W_t\}_{t=1}^T$ are drawn from the optimized variational distribution and the model's predictions are averaged:
\begin{equation}
\begin{split}
 p(y_{*} \rvert x_{*}, \mathcal{D}) &\approx \frac{1}{T} \sum_{t=1}^T p(y_{*} \rvert x_{*}, W_t), \\
 & \text{where } W_t \sim q_{\theta}(W)
\end{split}
 \label{eq:mc_approximation}
\end{equation}
The challenge lies in defining a variational distribution $q_{\theta}(W)$ that is \textbf{(1)} expressive enough to capture meaningful uncertainty and \textbf{(2)} simple enough to optimize via \cref{eq:elbo_objective} and sample from for \cref{eq:mc_approximation}.

\subsection{Monte Carlo Dropout (MCD)}
MCD~\cite{gal_dropout_2016} re-interprets the Dropout~\cite{srivastava_dropout_2014} regularizer as approximate Bayesian inference, showing that a neural network trained with dropout and L2 regularization implicitly optimizes the ELBO (\cref{eq:elbo_objective}).

In this framework, the variational distribution $q_{\theta}(W)$ is defined by the dropout mechanism itself. For a single weight matrix $W^l$ in layer $l$, the variational distribution $q(W^l)$ is defined over a matrix of learnable parameters $M^l$ (the weights $\theta$) and a random binary mask $z^l$:
\begin{equation}
 W^l = M^l \cdot \text{diag}(z^l), \quad \text{where } z^l_j \sim \text{Bernoulli}(p)
 \label{eq:mcd_q}
\end{equation}
Here, $p$ is the dropout probability. Critically, sampling from this $q_{\theta}(W)$ is trivial: it is simply a standard forward pass with dropout enabled.

Thus, MCD applies this insight at inference time. By performing $T$ stochastic forward passes (with dropout enabled) and averaging the results, one is directly computing the Monte Carlo approximation of the predictive posterior (\cref{eq:mc_approximation}). This provides a theoretically-grounded method for uncertainty quantification from a single model.

\subsection{Monte Carlo DropBlock (MCDB)}
While MCD provides a powerful framework, standard dropout is known to be less effective for convolutional layers, as spatial correlations limit its regularizing effect. To address this, DropBlock (DB)~\cite{ghiasi_dropblock_2018} was proposed as a structured regularizer that drops contiguous feature map regions.

MCDB~\cite{yelleni_monte_2024} extends the variational formalism of MCD to DB. Applying DB to a feature map $A^l$ is equivalent to imposing a structured variational distribution on the convolutional weights $W^{l+1}$.

In this formalism, $q(\tilde{W}^l)$ is defined as a product of Bernoulli distributions over \textit{blocks} of the weight tensor. Sampling from $q(\tilde{W}^l)$ is equivalent to either zeroing out a block of weights or keeping them set to learned values $W^l$:
\begin{equation}
 q(W^l_i) = W^l \cdot z^l_i, \quad \text{where } z^l_i \sim \text{Bernoulli}(\gamma)
 \label{eq:mcdb_q}
\end{equation}
Here, $i$ indexes a block in the filter and $\gamma$ is the DB probability. Similar to MCD, performing $T$ stochastic forward passes with DB enabled at inference time constitutes a valid MC approximation (\cref{eq:mc_approximation}) of the predictive posterior.

%% file: sec/4_MCSD.tex
\section{Monte Carlo Stochastic Depth}
\label{sec:MCSD}

Following the framework established in \cref{sec:background}, we now extend the paradigm of scalable Bayesian approximation to SD. We first formalize the SD mechanism and then demonstrate its connection to variational inference, thereby providing the theoretical grounding for the method we term \textbf{Monte Carlo Stochastic Depth (MCSD)} as a principled UQ method.

\subsection{Formalism for Stochastic Depth}
SD is a training regularizer designed explicitly for deep residual networks~\cite{huang_deep_2016}. A standard residual block $l$ (with, for simplicity, a skip connection) defines the transformation of its input $x_l$ to its output $x_{l+1}$ as:
\begin{equation}
 x_{l+1} = x_l + \mathcal{F}_l(x_l; W_l)
 \label{eq:res_block}
\end{equation}
where $W_l$ are the parameters of the residual function $\mathcal{F}_l$.

SD introduces a Bernoulli random variable $b_l \sim \text{Bernoulli}(p_l)$ for each block $l$, where $p_l$ is the "survival probability" of the block. The forward pass is modified to:
\begin{equation}
 x_{l+1} = x_l + b_l \cdot \mathcal{F}_l(x_l; W_l)
 \label{eq:sd_forward}
\end{equation}
During training, if $b_l=1$ (with probability $p_l$), the block is active. If $b_l=0$ (with probability $1-p_l$), the entire residual function $\mathcal{F}_l$ is "dropped," and the output is simply the identity: $x_{l+1} = x_l$. This effectively shortens the network, as only the identity path is active. At inference time, the standard SD protocol deactivates this stochasticity and uses a deterministic, scaled output: $x_{l+1} = x_l + p_l \cdot \mathcal{F}_l(x_l; W_l)$.

\subsection{Stochastic Depth as Approximate Variational Inference}
We posit that, analogous to MCD and MCDB, a network trained with SD is implicitly performing approximate variational inference. The key is to define the variational distribution $q_{\theta}(W)$ that this mechanism represents.

Unlike MCD/MCDB, which place a distribution over individual weights or small weight blocks, SD places a distribution over the inclusion of entire network stages.

Let the full set of learnable parameters be $\theta = \{W_l\}_{l=1}^L$ for all $L$ residual blocks. We define our variational distribution $q_{\theta}(W)$ as a discrete distribution over the network parameters. Specifically, $q_{\theta}(W)$ places mass on a finite collection of parameter values by stochastically zeroing out the weights of entire residual blocks, effectively simulating the selection of a sub-network. This process is governed by a vector of $L$ independent Bernoulli variables, $B = \{b_1, \dots, b_L\}$, where $b_l \sim \text{Bernoulli}(p_l)$.

A sample $W^{(B)} \sim q_{\theta}(W)$ is a specific sub-network configuration defined by the sampled vector $B$. The model's computation for this sample, $f(x \mid W^{(B)})$, is performed by applying \cref{eq:sd_forward} at each block.
\begin{equation}
 q_{\theta}(W) \equiv p(B) = \prod_{l=1}^L p(b_l) = \prod_{l=1}^L p_l^{b_l} (1-p_l)^{1-b_l}
 \label{eq:mcsd_q}
\end{equation}
The learnable parameters $\theta = \{W_l\}$ are the parameters of this variational distribution, and sampling from $q_{\theta}(W)$ is trivially achieved by performing a standard SD forward pass.

\subsection{Optimizing the ELBO}
The standard SD training objective can be interpreted as a practical surrogate for optimizing the ELBO under this variational distribution, rather than a strict analytical maximization. The first term of the ELBO (\cref{eq:elbo_objective}) is the expected log-likelihood. For a dataset $\mathcal{D}$ and a loss function $\mathcal{L}$ (e.g., cross-entropy), this evaluates to:
\begin{equation}
\begin{split}
  &\mathbb{E}_{B \sim q_{\theta}}[\sum_{(x,y) \in \mathcal{D}} \log p(y \rvert x, W^{(B)})] \\
  &\approx \sum_{(x,y) \in \mathcal{D}} \mathcal{L}(y, f(x \mid W^{(B)}))
\end{split}
\end{equation}
This expectation is approximated using Monte Carlo sampling. The standard SD training procedure, which samples one path $B$ per forward pass (or mini-batch) and computes the loss, is a stochastic gradient ascent optimization of this first ELBO term.

The second ELBO term (\cref{eq:elbo_objective}) is the complexity penalty. As with MCD \cite{gal_dropout_2016}, we assume a simple Gaussian prior $p(W)$ over the learnable parameters $\theta = \{W_l\}$. We acknowledge that rigorously computing the KL divergence between a discrete mixture distribution and a continuous prior is ill-posed. Following established methodologies for approximate Bayesian inference in deep learning, this KL divergence term is not optimized analytically but is instead approximated by the standard $L_2$ regularization (weight decay) applied to the weights $W_l$ during training.

Thus, the standard Stochastic Depth training procedure (stochastic forward pass + $L_2$ regularization) is shown to be a practical proxy for the optimization of the ELBO for the proposed variational distribution $q_{\theta}(W)$.

\subsection{Prediction and Uncertainty with MCSD}
The established connection to variational inference provides the theoretical grounding for applying this framework to Bayesian prediction. We follow the Monte Carlo approximation of the predictive posterior (\cref{eq:mc_approximation}) from \cref{sec:background}.

We define MCSD as the approximation of the predictive posterior by performing $T$ stochastic forward passes at inference time, each with a new, independently sampled Bernoulli vector $B_t$:
\begin{equation}
% p(y_{*} \rvert x_{*}, \mathcal{D}) \approx \frac{1}{T} \sum_{t=1}^T p(y_{*} \rvert x_{*}, W^{(B_t)}), \quad \text{where } B_t \sim q_{\theta}(W)
\begin{split}
 p(y_{*} \rvert x_{*}, \mathcal{D}) &\approx \frac{1}{T} \sum_{t=1}^T p(y_{*} \rvert x_{*}, W^{(B_t)}), \\
 & \text{where } B_t \sim q_{\theta}(W)
\end{split}
 \label{eq:mcsd_prediction}
\end{equation}
This discards the standard deterministic SD inference rule and instead utilizes the stochasticity to sample from the approximate posterior. Resulting metrics over the $T$ resulting predictions $p(y_{*} \rvert x_{*}, W^{(B_t)})$ provide a robust, theoretically-grounded measure of epistemic uncertainty. This formalism provides a theoretical interpretation for using SD as a Bayesian approximation and positions MCSD as a valid peer to MCD and MCDB.

\subsection{Algorithmic Formulation}

The theoretical framework culminating in \cref{eq:mcsd_prediction} translates directly into a practical algorithm. We formalize this practical implementation in \cref{alg:mcsd}. The core requirement is the retention of stochastic path sampling during the inference phase, enabled by a direct implementation of the Monte Carlo sampling step (\cref{eq:mcsd_prediction}).

\begin{algorithm}[H]
\small
\caption{Monte Carlo Stochastic Depth}
\label{alg:mcsd}
\begin{algorithmic}[1]
\Statex \hspace*{-\algorithmicindent} \textbf{Input:} Feature representation from a residual path ($A_{res}$), identity shortcut ($x$)
\Statex \hspace*{-\algorithmicindent} \textbf{Parameters:} mode, $p_{drop}$
\If{mode == training or mode == inference}
    \State Set survival probability: $p_{keep}=1.0-p_{drop}$
    \State Get batch size $B$ from $A_{res}$
    \State Randomly sample a per-sample mask $M: M_b \sim \text{Bernoulli}(p_{keep})$ for $b\in[1,\dots,B]$
    \State Reshape $M$ to broadcast across $A_{res}$ (e.g., shape $[B,1,1,1]$)
    \State Apply the mask: $A_{res}=A_{res}\times M$
    \State Normalize the features: $A_{res}=A_{res}/p_{keep}$
\EndIf
\State \textbf{Return} $x+A_{res}$
\end{algorithmic}
\end{algorithm}

The per-sample mask $M$ (Line 4) is the practical realization of the Bernoulli variable $b_l$ (\cref{eq:sd_forward}), and normalization (Line 8) maintains output consistency. It is critical to note that MCSD inherently requires skip-connections to function as a Bayesian approximation, as it relies on stochastically nullifying the residual function $\mathcal{F}_l$ (realized as $A_{res}$) while preserving the identity path (Line 9). This makes it a natural fit for the residual-based backbones of modern detectors.

%% file: sec/5_experiments.tex
\section{Experiments}
\label{sec:experiments}

\begin{figure*}[t]
  \centering
   \includegraphics[width=\textwidth]{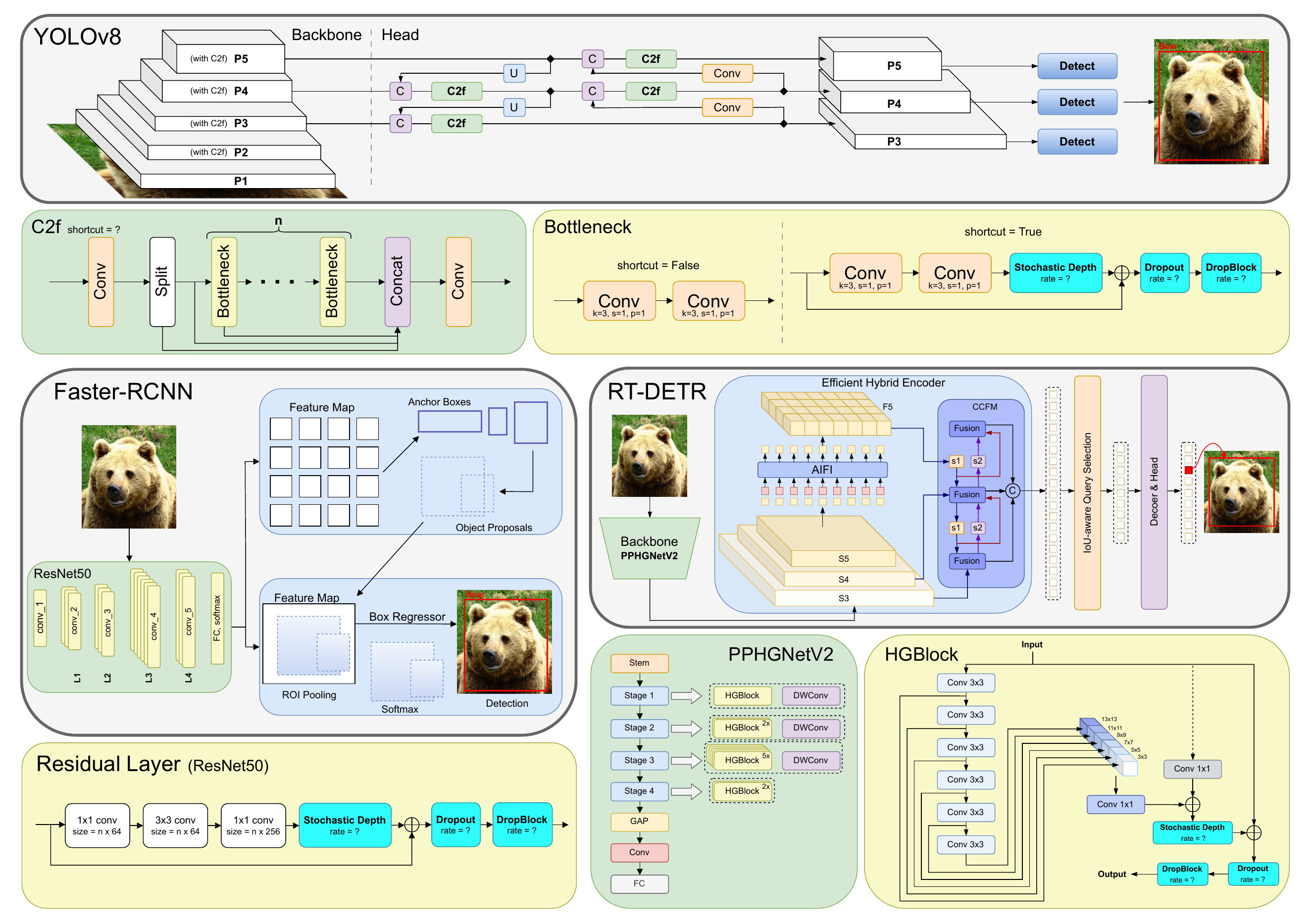}
   \caption{Architectural overview of the detectors used in our study: YOLOv8, Faster R-CNN, and RT-DETR. The expanded block diagrams (e.g., Bottleneck, Residual Layer, HGBlock) illustrate the specific insertion points within the residual paths where MCD, MCDB, and MCSD are applied.}
   \label{fig:implementationDetails}
\end{figure*}

To validate our theoretical derivation and assess the practical utility of MCSD, we present a comprehensive empirical study. We benchmark its performance against prominent stochastic UQ methods on modern object detection architectures. Our evaluation focuses on three main objectives:

\begin{enumerate}[label=(\roman*)]
    \item \textbf{Effectiveness in Predictive Performance:} Assess MCSD for uncertainty estimation and calibration against the prevalent methods MCD and MCDB.
    \item \textbf{Performance in Distribution Shift Scenarios:} Examine the efficacy of UQ on distributionally shifted datasets for MCSD. 
    \item \textbf{Adaptability across Architectures and Implementation Parameters:} Demonstrate the applicability of MCSD across different DNN architectures.
\end{enumerate}

\subsection{Experimental Setup}
\label{subsec:ex_setup}

\noindent\textbf{Models.} We conduct experiments on a diverse set of state-of-the-art object detectors to investigate the broad applicability of MCSD. This includes Faster R-CNN~\cite{ren_faster_2015} (CNN-based, two-stage), YOLOv8x~\cite{Jocher_Ultralytics_YOLO_2023} (CNN-based, single-stage), and RT-DETRx~\cite{zhao_detrs_2024} (Transformer-based, single-stage).

\noindent\textbf{Baselines.} We compare MCSD against the two most prominent stochastic regularizer-based UQ methods: Monte Carlo Dropout (MCD)~\cite{gal_dropout_2016} and Monte Carlo DropBlock (MCDB)~\cite{yelleni_monte_2024}.

\noindent\textbf{Datasets.} We use the large-scale COCO dataset (Common Objects in Context)~\cite{lin_microsoft_2014}. We follow standard literature practices, training on the \texttt{train2017} split and using \texttt{val2017} as our in-distribution (ID) test set. To evaluate robustness and uncertainty quality under distribution shift, we use the COCO-O (COCO-Out-of-distribution) dataset~\cite{mao_coco-o_2023}.

\noindent\textbf{Evaluation Metrics.} We evaluate all methods across three key dimensions: predictive accuracy, uncertainty calibration, and uncertainty ranking.

We report the standard COCO metric, mean Average Precision (mAP), computed at an IoU threshold of 0.5:0.95. This assesses the core detection performance of the models.

To quantify predictive uncertainty, we require a scalar score $U(\mathbf{x})$ for each prediction. To consolidate the raw detections across the $T$ stochastic forward passes into individual object observations, we employ the Basic Sequential Algorithmic Scheme (BSAS) with spatial and semantic affinity~\cite{miller_evaluating_2019}. Given the architectural differences, we adapt our uncertainty metric as follows:
\begin{itemize}
    \item \textbf{For Faster R-CNN (multi-class):} The model uses a softmax output for class predictions. We compute the Shannon Entropy~\cite{shannon_mathematical_1948} over the mean predictive probability distribution $\mathbf{\bar{p}} = \frac{1}{T}\sum_{t=1}^T \mathbf{p}_t$ for the $C$ object classes.% (plus background):
    \item \textbf{For YOLO \& RT-DETR (multi-label):} These models use per-class sigmoid activations, treating detection as a multi-label problem. We first compute the mean predictive probability $\bar{p}_c = \frac{1}{T}\sum_{t=1}^T p_{t,c}$ for each class $c$. The total uncertainty is the mean of binary entropies across all $C$ classes.
\end{itemize}

\noindent For uncertainty calibration we evaluate how well the model's predictive confidence aligns with its empirical accuracy:
\begin{itemize}
    \item \textbf{Brier Score (BS)}~\cite{brier1950verification}\textbf{:} Measures the mean squared error between the predicted probabilities and the one-hot ground-truth labels $\mathbf{y}$. A lower Brier Score indicates better calibration.
    \item \textbf{Expected Calibration Error (ECE)}~\cite{guo_calibration_2017}\textbf{:} This metric bins predictions based on their confidence score (e.g., $\max_c \bar{p}_c$) and computes the weighted average difference between accuracy and confidence within each bin. A lower ECE indicates better calibration.
\end{itemize}

\noindent Uncertainty ranking is shown via the \textbf{Area Under the Accuracy-Rejection Curve (AUARC)}~\cite{geifman_bias-reduced_2019}. The accuracy-rejection curve plots accuracy against the fraction of data rejected $\theta \in [0, 1]$, by progressively rejecting the most uncertain predictions. A perfect uncertainty score would reject all incorrect predictions before any correct ones, where a higher AUARC shows a more reliable uncertainty ranking.

\subsubsection{Implementation Details}

\noindent\textbf{Model Implementations.} Our models are based on standard public implementations: Torchvision~\cite{paszke_pytorch_2019} for Faster R-CNN, and Ultralytics~\cite{Jocher_Ultralytics_YOLO_2023} for YOLOv8x and RT-DETRx. We integrated the stochastic methods (MCD, MCDB, MCSD) by modifying the primary skip-connection blocks within each architecture (\cref{fig:implementationDetails}). For Faster R-CNN, we utilized the timm~\cite{Wightman_PyTorch_Image_Models} library for DropBlock and DropPath. For the Ultralytics models, we designed custom Bottleneck (YOLOv8x) and HGBlock (RT-DETRx) modules incorporating DropBlock based on the work of Yelleni \etal~\cite{yelleni_monte_2024} and Stochastic Depth as detailed in \cref{alg:mcsd} (\cref{sec:MCSD}).

\noindent\textbf{Training and Adaptation.} Initialized with COCO pre-trained weights, models were finetuned with active stochastic layers. Faster R-CNN was finetuned end-to-end for 12 epochs (learning rate $5 \times 10^{-3}$). Conversely, YOLOv8x and RT-DETRx utilized parameter-efficient finetuning, updating only the modified stochastic layers for 20 epochs (learning rate $1 \times 10^{-4}$). This extended finetuning duration was critical; minimal finetuning made the models fragile, necessitating impractically low drop rates ($\approx$0.01). Finally, we modified both frameworks to extract the full class probability vectors, which are not natively returned (see~\cref{SupplMat:FramewAdapts}).

\noindent\textbf{Ablation Parameters.} To ensure a rigorous comparison, we conduct extensive ablation studies. We vary the drop rate (0.01 to 0.25), the number of Monte Carlo samples $T$ (5 to 20), and the prediction confidence threshold (0.05 to 0.7). Furthermore, we analyze the impact of where stochasticity is introduced by varying the set of adapted blocks (e.g., all blocks, only early blocks, or only late blocks).

\begin{figure}[t]
  \centering
   \includegraphics[width=\linewidth]{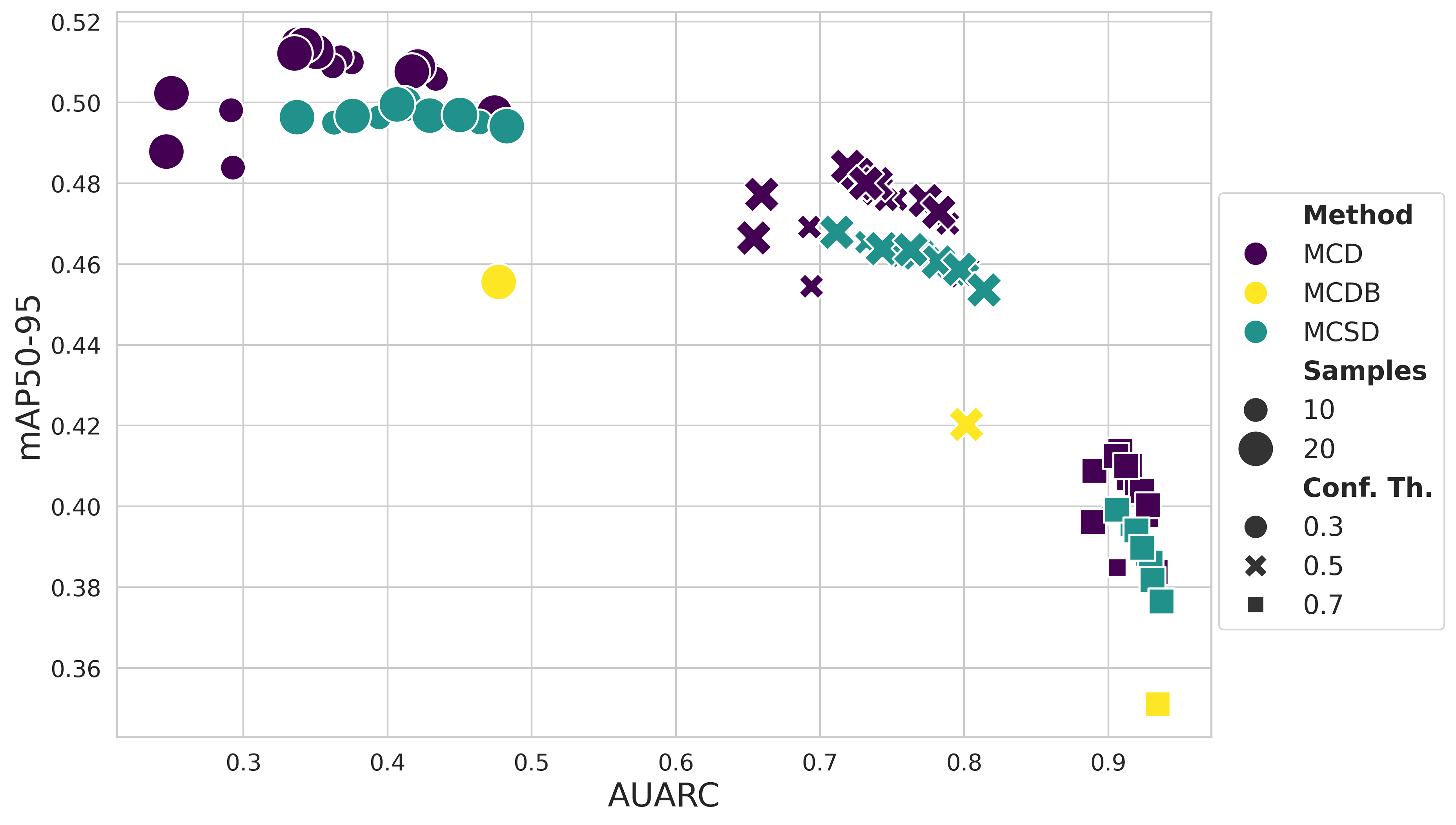}
   \caption{Pareto front analysis of the accuracy (mAP) vs. uncertainty ranking (AUARC) trade-off for RT-DETRx on the COCO validation set. Each point represents a unique hyperparameter configuration. Non-competitive MCDB configurations (mAP $< 0.1$) are omitted.}
   \label{fig:ParetoTradeof}
\end{figure}

\subsection{Results}

\textbf{Accuracy-Uncertainty Trade-off.} \Cref{fig:ParetoTradeof} illustrates a Pareto front for the trade-off between predictive performance (mAP) and uncertainty ranking (AUARC). This plot is generated from a sweep of hyperparameter configurations for RT-DETRx; we omit YOLOv8x and Faster R-CNN for clarity, as they demonstrated similar trends.

First, MCSD and MCD demonstrate a similar Pareto frontier, with MCD often holding a slight advantage in mAP and MCSD showing a slight advantage in AUARC. Second, MCDB proved highly sensitive in YOLOv8x and RT-DETRx; applying DropBlock anywhere but the final backbone layer (see \cref{fig:implementationDetails}) resulted in severe performance degradation and were thus omitted as non-competitive. This fragility likely accounts for MCDB's slightly lower performance in our configurations, suggesting its optimal placement is highly architecture-specific and may lie outside the backbone entirely, such as closer to the final detection head, as was successfully demonstrated by Yelleni, S.H. \etal~\cite{yelleni_monte_2024}. Finally, while the Monte Carlo sample size ($T$) has a minor effect, the confidence threshold is the dominant tuning parameter for navigating the mAP-AUARC balance.

\begin{table*}[t]
\centering
\scriptsize
\caption{Quantitative comparison of the best Pareto-optimal configurations on COCO. This table isolates the top-performing setup for each method, evaluating accuracy (mAP), calibration (Brier, ECE), and uncertainty ranking (AUARC). Bold values highlight the best method for each metric and model.}
\label{tab:id_results}
\begin{tabular}{lllccccccc}
\toprule
\textbf{Model} & \textbf{Method} & \textbf{Adapted Layers} & \textbf{Drop Rate} & \textbf{MC-Samples} & \textbf{Conf. Thresh.} & {\textbf{mAP.5.95} $\uparrow$} & {\textbf{Brier $\times 10^{-3}$} $\downarrow$} & {\textbf{ECE} $\downarrow$} & {\textbf{AUARC} $\uparrow$} \\
\midrule
Faster R-CNN & MCD  & All layers & 0.20 & 20 & 0.40 & \textbf{0.270} & 5.36 & 0.213 & 0.680 \\
Faster R-CNN & MCDB & All layers & 0.20 & 20 & 0.40 & 0.258 & 5.22 & \textbf{0.200} & 0.653 \\
Faster R-CNN & MCSD & Layer 4 & 0.10 & 20 & 0.50 & 0.266 & 5.97 & 0.247 & \textbf{0.741} \\
\midrule
YOLOv8x & MCD  & Last 15 layers & 0.20 & 20 & 0.15 & \textbf{0.505} & 2.43 & 0.060 & 0.668 \\
YOLOv8x & MCDB & Last layer & 0.05 & 10 & 0.15 & 0.473 & 2.08 & 0.063 & 0.771 \\
YOLOv8x & MCSD & First 15 layers & 0.05 & 20 & 0.20 & 0.496 & \textbf{1.93} & \textbf{0.035} & \textbf{0.778} \\
\midrule
RT-DETRx & MCD  & Last six layers & 0.20 & 20 & 0.50 & \textbf{0.484} & \textbf{5.07} & 0.088 & 0.726 \\
RT-DETRx & MCDB & Last layer & 0.20 & 20 & 0.50 & 0.420 & 6.64 & 0.077 & \textbf{0.813} \\
RT-DETRx & MCSD & Last six layers & 0.20 & 20 & 0.50 & 0.463 & 6.08 & \textbf{0.073} & 0.784 \\
\bottomrule
\end{tabular}
\end{table*}

\vspace{1mm}
\noindent \textbf{Quantitative In-Distribution Results.} \Cref{tab:id_results} isolates the Pareto-optimal configuration (best mAP/AUARC balance) for each model-method combination. The table confirms the trends from the Pareto plot: MCD consistently yields the highest mAP, while MCSD generally delivers superior uncertainty quantification, achieving the best AUARC or ECE in the majority of configurations.

Contrasting prior related work~\cite{Yao_SDinVI_2024} noting improvements in both accuracy and ECE for segmentation, our detection benchmark reveals a clearer trade-off: MCD excels in predictive accuracy, whereas MCSD provides more robust uncertainty and calibration. Faster R-CNN, our weakest mAP baseline, produced the poorest calibration scores, suggesting a link between high overconfidence (high ECE) and lower predictive performance. Architecturally, the finer weight-level perturbations in MCD preserve baseline accuracy whereas the coarser dropping of entire residual paths by MCSD generates the information variance required for superior uncertainty ranking.

\vspace{1mm}
\noindent \textbf{Ablation on Stochastic Layer Placement.} Furthermore, our study reveals that the optimal placement of the stochastic layers is a critical, architecture-dependent hyperparameter. We observe distinct behaviors across the models:
\begin{itemize}
    \item \textbf{Faster R-CNN:} For MCD and MCSD, applying stochasticity to earlier layers yielded improvements in calibration (ECE) and ranking (AUARC). Conversely, MCDB performed best when applied to later layers.
    \item \textbf{YOLOv8x:} This model showed a clear divergence. Stochasticity in earlier layers produced optimal calibration (ECE), while placement in later layers optimized uncertainty ranking (AUARC). MCDB was non-viable in earlier layers, causing a significant drop in performance. %degradation. %degradation.
    \item \textbf{RT-DETRx:} In contrast to the other models, all three methods (MCD, MCDB, and MCSD) achieved their best ECE and AUARC scores when stochasticity was applied to the later layers of the network.
\end{itemize}

\begin{figure}[t]
  \centering
   \includegraphics[width=\linewidth]{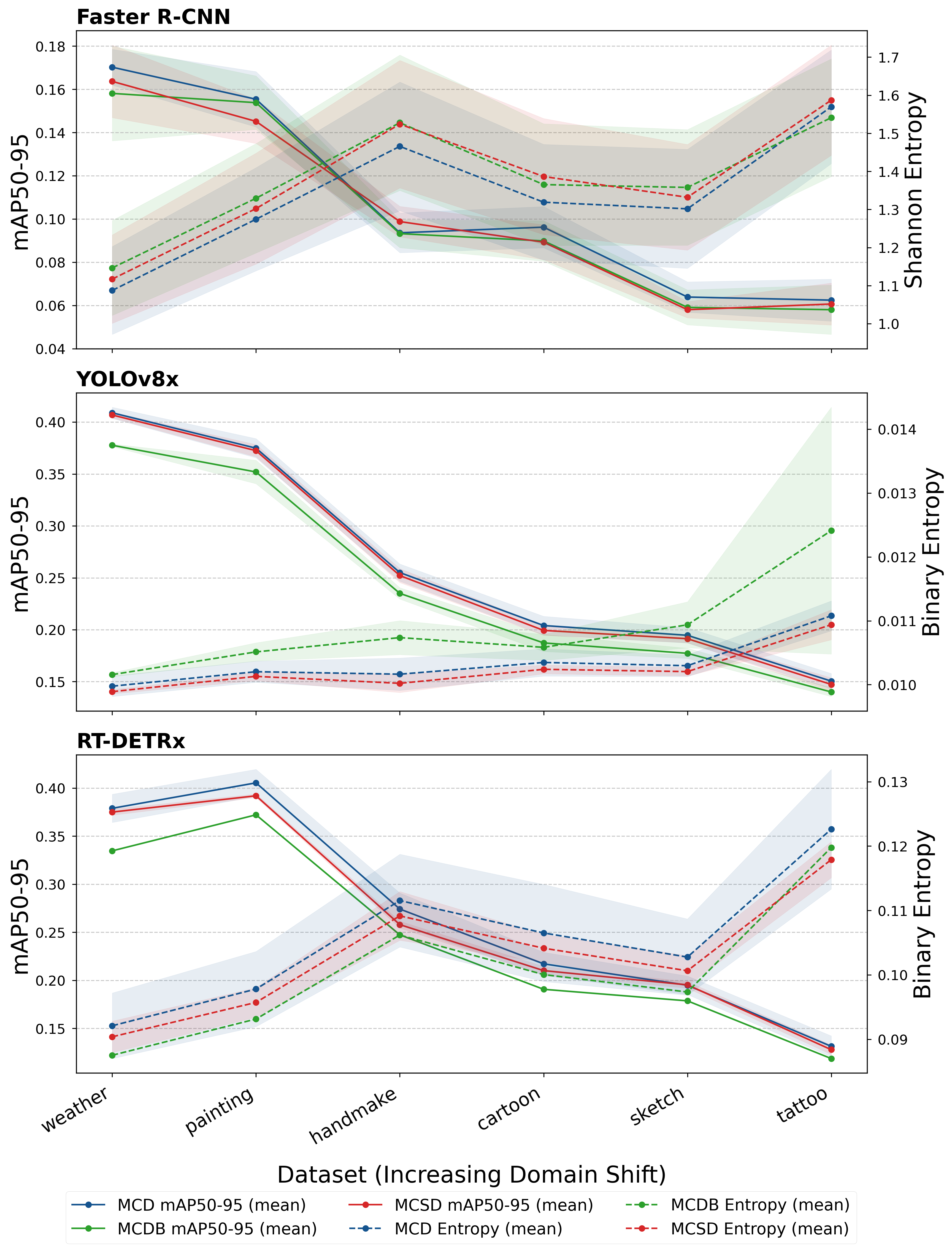}
   \caption{Performance under domain shift on the COCO-O benchmark for Faster-RCNN (top), YOLOv8x (middle), and RT-DETRx (bottom). Solid lines track mAP, dashed lines predictive entropy.}
   \label{fig:ood_shift}
\end{figure}

\subsubsection{Distribution Shift Analysis}

To evaluate performance on OOD data, we use the COCO-O benchmark, which introduces increasing stylistic and contextual shifts such as weather variations, cartoons, and sketches. \Cref{fig:ood_shift} plots the resulting mAP degradation against the change in predictive entropy to measure how well each method's uncertainty reflects the increasing shift.

As \cref{fig:ood_shift} illustrates, all methods exhibit a significant drop in mAP as the domain shift intensifies (e.g., from 'weather' to 'tattoo'). While this sharp accuracy degradation is met with only a slight increase in predictive entropy, a consistent upward trend is nonetheless observable for all methods, indicating sensitivity to the domain shift. Crucially, the performance of MCSD under these shifts, both in mAP and entropy response, is nearly identical to that of MCD. This parity demonstrates that MCSD is not only a peer to MCD on in-distribution data but also an equally robust alternative under significant domain shift, reinforcing its viability as a practical UQ method.

%% file: sec/6_limitations.tex
\section{Limitations}
\label{sec:limitations}

While MCSD improves relative uncertainty ranking, absolute calibration remains sensitive to the underlying model architecture. As detailed in the supplementary material~\cref{tab:ood_performance}, dense detectors (YOLOv8x, RT-DETRx) exhibit a compressed entropy range under severe domain shift compared to the two-stage model Faster R-CNN, hinting at intrinsic overconfidence. Furthermore, evaluating uncertainty in object detection is inherently sensitive to the confidence thresholds used to filter bounding boxes prior to metric calculation. Future work should explore standardized, threshold-free evaluation protocols to decouple uncertainty quantification from standard detection filtering. Additional structural constraints are discussed in~\cref{SupplMat:Limitations}.

%% file: sec/7_conclusion.tex
\section{Conclusion}
\label{sec:conclusion}

In this work, we establish Monte Carlo Stochastic Depth (MCSD) as a principled approximate Bayesian method, grounding its use for UQ. We validated this theory with the first comprehensive empirical benchmark of MCSD for the task of object detection, evaluating it against MCD and MCDB on state-of-the-art architectures (YOLO, RT-DETR) using the COCO and COCO-O datasets.

Our results reveal a compelling trade-off: while MCD often maintains a slight edge in predictive accuracy (mAP), MCSD achieves superior uncertainty quality in calibration (ECE) and uncertainty ranking (AUARC). We establish MCSD as a theoretically-grounded, and empirically-validated tool, positioning it as a strong alternative, especially in safety-critical systems where reliable calibration and uncertainty-ranking are paramount.

%% file: sec/A_sup_Reproducibility.tex
\clearpage
\setcounter{page}{1}
\maketitlesupplementary

\section{Appendix A: Reproducibility}
\label{sec:app-reproduc}

Complete source code reproducing all methods (MCD, MCDB, MCSD) and models (Faster R-CNN, YOLOv8x, RT-DETRx) is provided in the \href{https://github.com/code-supplement-2026/mc-val}{GitHub}\footnote{https://github.com/code-supplement-2026/mc-val} repository.

\noindent\textbf{Hardware and Software.} Experiments were conducted using Python 3.10.12, PyTorch 2.6.0, and Ultralytics 8.3.171. The codebase is hardware-agnostic; however, all reported results were generated on a single NVIDIA GeForce RTX 3090 GPU. Run-times may vary across different hardware configurations. Installation guides and Quick Start instructions are included in the repository, with dependencies listed in \texttt{requirements.txt}.

\subsection{Framework Adaptations} \label{SupplMat:FramewAdapts}
\noindent \textbf{Network Architectures.} Integrating stochastic regularization into the standard architectures of Faster R-CNN, YOLOv8x, and RT-DETRx (as visualized in \cref{fig:implementationDetails}) necessitated specific modifications to the underlying framework definitions.

\begin{itemize}
    \item \textbf{Faster R-CNN:} We extended the \texttt{Bottleneck} class within Torchvision's \texttt{resnet.py}. To incorporate DropBlock and Stochastic Depth, we leveraged existing implementations from the \texttt{timm}~\cite{Wightman_PyTorch_Image_Models} library. This custom \texttt{Bottleneck} was then integrated into the backbone instantiation via the \texttt{fasterrcnn\textunderscore resnet50\textunderscore fpn} function in \texttt{faster\textunderscore rcnn.py}.

    \item \textbf{YOLOv8x / RT-DETRx:} For the Ultralytics-based networks, we modified \texttt{nn/modules/block.py} to support the proposed stochastic methods. We integrated the MCDB implementation by Yelleni \etal~\cite{yelleni_monte_2024} and the MCSD logic defined in \cref{alg:mcsd}. These stochastic mechanisms were injected directly into the \texttt{Bottleneck} class for YOLOv8x and the \texttt{HGBlock} class for RT-DETRx. For MCD, we utilized the framework's native Dropout modules.
\end{itemize}

\vspace{2mm}
\noindent \textbf{Inference Outputs.} To compute the entropy-based uncertainty scores described in \cref{subsec:ex_setup}, access to the full class probability vectors, which are typically discarded during post-processing, was required. We adapted the inference pipelines of both frameworks as follows:

\begin{itemize}
    \item \textbf{Ultralytics:} We modified the inference flow to retain the full class probability vector prior to thresholding. This involved updating the \texttt{postprocess} method within the \texttt{DetectionPredictor} class (found in the model-specific \texttt{predict.py} files), the \texttt{BasePredictor} class in \texttt{engine/predictor.py}, and the \texttt{non\textunderscore max\textunderscore suppression} logic in \texttt{utils/ops.py}.

    \item \textbf{Torchvision:} We modified the \texttt{postprocess\textunderscore detections} function within the \texttt{RoIHeads} class (located in \texttt{detection/roi\textunderscore heads.py}) to preserve and return the full probability tensors alongside the standard detection outputs.
\end{itemize}

\subsection{Uncertainty and Calibration Metrics}

We evaluated model calibration and predictive uncertainty using four primary metrics. Aligning with \cref{subsec:ex_setup} the formulations are detailed below.

\vspace{2mm}
\noindent \textbf{Predictive Entropy.}
Entropy measures the uncertainty inherent in a probability distribution. As some detectors (YOLOv8x, RT-DETRx) produce independent sigmoid probabilities $\mathbf{p}_i = (p_{i,1}, \dots, p_{i,C})$ for each class, where probabilities do not sum to 1, we distinguish between two formulations:

\begin{enumerate}[label=\arabic*.]
    \item \textbf{Mean Binary Entropy:} We compute the entropy for each class $c$ as an independent Bernoulli trial and then average these entropies. The binary entropy for a single class probability $p_{i,c}$ is:
    \begin{equation*}
        H(p_{i,c}) = -p_{i,c} \log_2(p_{i,c}) - (1-p_{i,c}) \log_2(1-p_{i,c})
    \end{equation*}
    The final metric for detection $i$, is the mean of these binary entropies:
    \begin{equation*}
        H_{\text{binary}}(\mathbf{p}_i) = \frac{1}{C} \sum_{c=1}^{C} H(p_{i,c})
    \end{equation*}

    \item \textbf{Shannon Entropy:} The standard Shannon entropy is used for models that output a single multinomial probability distribution where $\sum_{c=1}^{C} p_{i,c} = 1$ (i.e., a softmax output):
    \begin{equation*}
        H_{\text{shannon}}(\mathbf{p}_i) = - \sum_{c=1}^{C} p_{i,c} \log_2(p_{i,c})
    \end{equation*}
\end{enumerate}

\vspace{2mm}
\noindent \textbf{Predictive Uncertainty.}
We assessed the quality and practical utility of the predictive uncertainties. This was evaluated using metrics that measure calibration (Brier Score, ECE) and the effectiveness of the uncertainty score in identifying potential errors (AUARC).

\begin{itemize}
    \item \textbf{Brier Score (BS):} The Brier Score measures the accuracy of probabilistic predictions by computing the mean squared error (MSE) between the predicted probability vector and the one-hot encoded true label vector. Our implementation computes the average MSE over all $N$ detections and all $C$ classes.
    \begin{equation*}
        \text{BS} = \frac{1}{N \cdot C} \sum_{i=1}^{N} \sum_{c=1}^{C} (p_{i,c} - y_{i,c})^2
    \end{equation*}
    where $N$ is the total number of detections (True Positives in this context), $C$ is the number of classes, $p_{i,c}$ is the predicted probability for detection $i$ of class $c$, and $y_{i,c}$ is the one-hot true label ($1$ if $c$ is the true class for detection $i$, $0$ otherwise).

    \item \textbf{Expected Calibration Error (ECE):} ECE measures calibration by partitioning all $N$ detections (True Positives and False Positives) into $M$ equally-spaced confidence bins. It then computes a weighted average of the absolute difference between the mean accuracy and mean confidence in each bin.
    \begin{equation*}
        \text{ECE} = \sum_{m=1}^{M} \frac{|B_m|}{N} \left| \text{acc}(B_m) - \text{conf}(B_m) \right|
    \end{equation*}
    where $B_m$ is the set of detections whose confidence falls into bin $m$, $\text{acc}(B_m)$ is the accuracy (fraction of True Positives) of the detections in $B_m$, and $\text{conf}(B_m)$ is the average confidence of detections in $B_m$.

    \item \textbf{Area Under the Accuracy-Rejection Curve (AUARC):} AUARC evaluates the trade-off between predictive accuracy and the fraction of predictions rejected based on an uncertainty score. Following \cite{zeevi_ratein_2025}, detections are sorted by their uncertainty score (e.g., entropy) in descending order.
    \begin{equation*}
        \text{AUARC} = \int_{0}^{1} \text{Acc}(r) dr
    \end{equation*}
    where $r$ is the rejection fraction (from 0 to 1). $\text{Acc}(r)$ is the precision (i.e., $\frac{TP}{TP+FP}$) of the \textit{retained} (non-rejected) predictions after rejecting the top $r$ fraction of most uncertain detections. A higher AUARC value indicates that the uncertainty measure is a good proxy for error, allowing for the effective rejection of false positives.
\end{itemize}

%% file: sec/B_supp_theory.tex
\section{Appendix B: Theoretical Derivations}
\label{sec:app-theoretical}

In this section, we provide the formal derivation connecting the Stochastic Depth (SD) training objective to the Variational Inference (VI) framework utilized in the main paper.

\subsection{Derivation of the ELBO Objective for MCSD}
\label{app:elbo_derivation}

As defined in the main paper, our objective is to maximize the Evidence Lower Bound (ELBO):
\begin{equation*}
    \loss_{\text{VI}}(\params) = \underbrace{\mathbb{E}_{q_{\theta}(W)}[\log p(\data \rvert \weights)]}_{\text{Expected Log-Likelihood}} - \underbrace{\KL(q_{\params}(\weights) || p(\weights))}_{\text{Complexity Penalty}}
\end{equation*}
Here, we demonstrate that the standard SD training procedure constitutes a stochastic gradient optimization of this objective.

\subsubsection{The Expected Log-Likelihood Term}

Let the training dataset be $\data = \{(x_n, y_n)\}_{n=1}^N$. In the MCSD framework, the variational distribution $q_{\params}(\weights)$ represents the stochastic process of selecting a sub-network $\weights^{(\pathvec)}$ conditioned on a Bernoulli vector $\pathvec \in \{0,1\}^L$, as defined in Eq. (12) of the main text.

The expected log-likelihood term of $\loss_{\text{VI}}(\params)$ can be expanded as:
\begin{equation*}
    \mathbb{E}_{q_{\theta}(W)}[\log p(\data \rvert \weights)] = \mathbb{E}_{\pathvec \sim p(\pathvec)}\left[\sum_{n=1}^N \log p(y_n \rvert x_n, \weights^{(\pathvec)})\right]
\end{equation*}
Exact computation of this expectation requires summation over $2^L$ possible sub-networks, which is intractable. We approximate this via Monte Carlo sampling. For a mini-batch of size $M$, we sample a mask $\pathvec_m \sim p(\pathvec)$ and compute the estimator:
\begin{equation*}
        \mathbb{E}_{\pathvec \sim p(\pathvec)}\left[\sum_{n=1}^N \log p(..) \right] \approx \frac{N}{M} \sum_{m=1}^M \log p(y_m \rvert x_m, \weights^{(\pathvec_m)})
\end{equation*}
For standard discriminative tasks, the model's log-likelihood corresponds to the negative of the task loss function (e.g., Cross-Entropy or Focal Loss), such that $\log p(y \rvert x, \weights) \propto -\loss_{\text{task}}(y, f(x; \weights))$.

Consequently, maximizing the expected log-likelihood is equivalent to minimizing the expected task loss:
\begin{equation*}
    \begin{split}
        &\arg \max_{\params} \mathbb{E}_{q_{\theta}(W)}[\log p(\data \rvert \weights)] \\
        &\iff \arg \min_{\params} \mathbb{E}_{\pathvec \sim p(\pathvec)}[\loss_{\text{task}}(y, f(x \mid \weights^{(\pathvec)}))]
    \end{split}
\end{equation*}
The standard SD training algorithm, which performs a forward pass with a sampled path $\pathvec$ and minimizes $\loss_{\text{task}}$, is therefore a direct stochastic gradient ascent optimization of the first term of the ELBO.

\subsection{The KL Divergence as L2 Regularization}
\label{app:kl_l2}

Having established the optimization of the likelihood term, we address the complexity penalty, $\KL(q_{\params}(\weights) || p(\weights))$.

In our framework, $q_{\params}(\weights)$ is a discrete mixture distribution over sub-networks, while the prior $p(\weights)$ is assumed to be a standard Gaussian $\mathcal{N}(0, I)$. Direct computation of the KL divergence between a discrete mixture and a continuous prior is ill-posed.

Following the established methodology for approximate Bayesian inference in deep learning \cite{gal_dropout_2016}, we approximate the complexity penalty via $L_2$ regularization (weight decay). We assume a prior length-scale, such that the minimization of the KL divergence corresponds to minimizing the $L_2$ norm of the variational parameters (the weights $\weights_l$ of the residual blocks). 

The resulting surrogate training objective becomes:
\begin{equation*}
    \loss_{\text{final}} = \frac{1}{M} \sum_{m=1}^M \loss_{\text{task}}(y_m, f(x_m \mid \weights^{(\pathvec_m)})) + \lambda \sum_{l=1}^L ||\weights_l||_2^2
\end{equation*}
This confirms that training with Stochastic Depth and weight decay optimizes a valid proxy for the full ELBO, justifying MCSD as a theoretically grounded Bayesian approximation.

\subsection{Justification for Probabilistic Scaling}
\label{app:scaling}

\cref{alg:mcsd} details the implementation of the MCSD residual block. A critical component is the scaling of the residual features $A_{res}$ by the inverse survival probability $1/p_l$ during the forward pass.

This scaling ensures that the stochastic gradients computed during training (and the predictions during MCSD inference) act as unbiased estimators of the full network. Let $x_{l+1}$ be the output of block $l$. The expected output over the distribution of masks $b_l$ is:
\begin{align*}
    \mathbb{E}_{b_l}[x_{l+1}] &= \mathbb{E}_{b_l}\left[x_l + \frac{b_l}{p_l} \cdot \mathcal{F}_l(x_l; \weights_l)\right] \nonumber \\
    &= x_l + \frac{\E[b_l]}{p_l} \cdot \mathcal{F}_l(x_l; \weights_l) \nonumber \\
    &= x_l + \mathcal{F}_l(x_l; \weights_l)
\end{align*}
Since $\E[b_l] = p_l$, the term cancels out, ensuring the expected output of the stochastic pass equals the output of the deterministic, fully active block. This preserves the feature magnitude distribution between training and Monte Carlo inference, preventing distributional shift when sampling $T$ predictions for uncertainty quantification.

%% file: sec/CD_sup_AddResultsDiscussion.tex
\section{Appendix C: Additional Results}
\label{sec:app-results}

\begin{figure}[h]
  \centering
   \includegraphics[width=\linewidth]{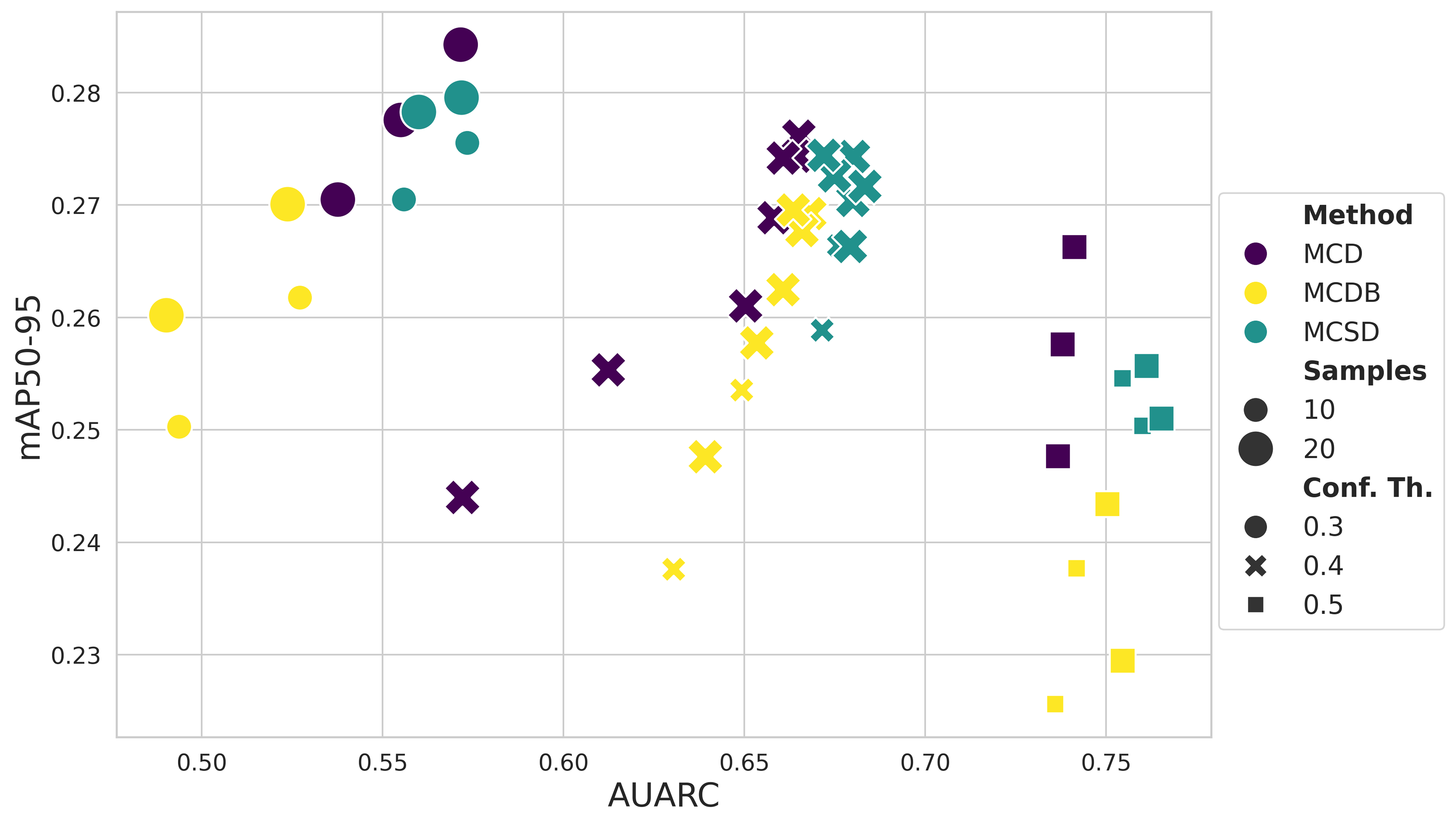}
   \caption{Pareto front analysis of the accuracy (mAP) vs. uncertainty ranking (AUARC) trade-off for Faster R-CNN on the COCO validation set. Each point represents a unique hyperparameter configuration. Non-competitive configurations are omitted.}
   \label{fig:ParetoTradeofFasterRCNN}
\end{figure}

\begin{figure}[h]
  \centering
   \includegraphics[width=\linewidth]{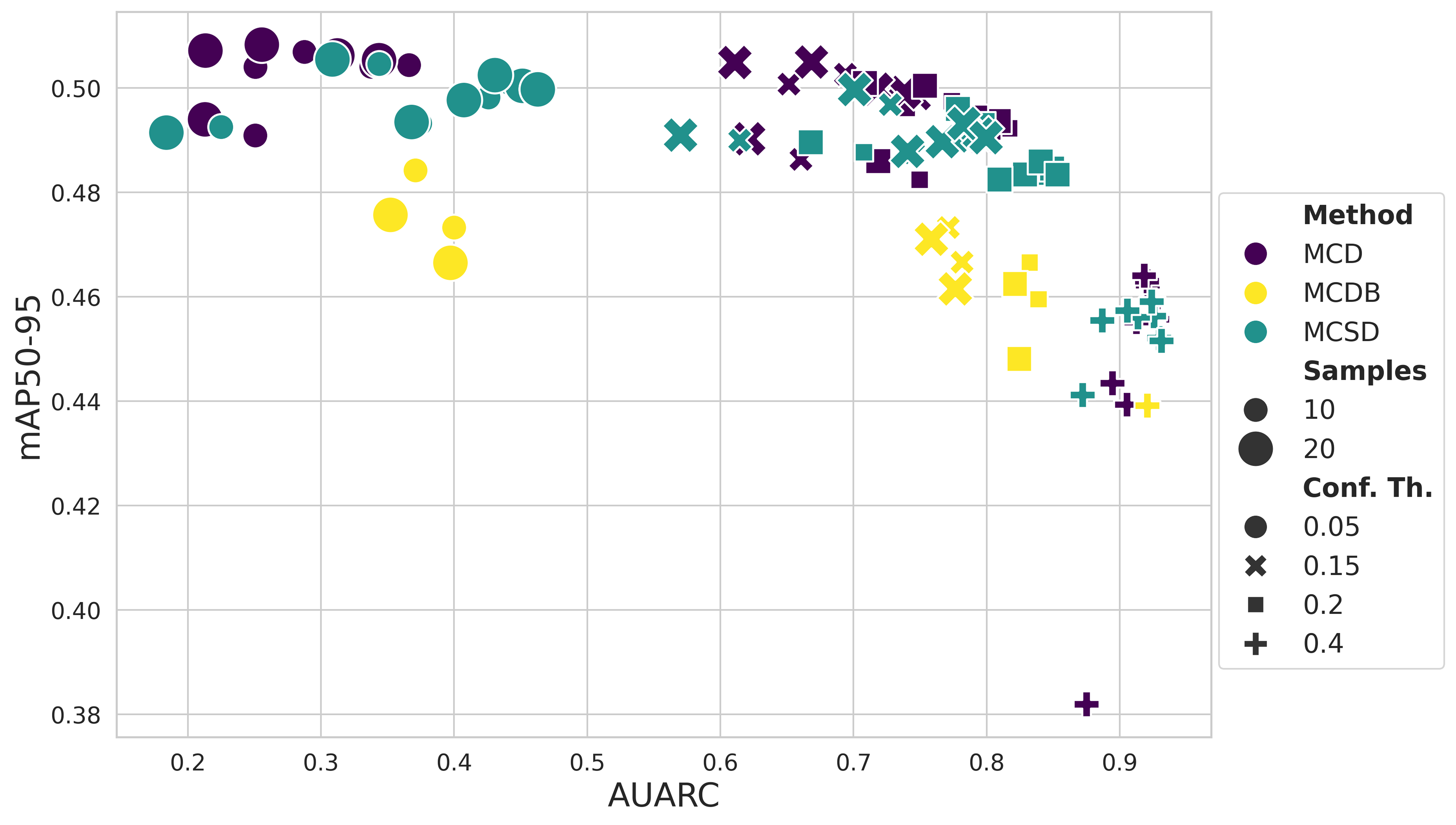}
   \caption{Pareto front analysis of the accuracy (mAP) vs. uncertainty ranking (AUARC) trade-off for YOLOv8x on the COCO validation set. Each point represents a unique hyperparameter configuration. Non-competitive configurations are omitted.}
   \label{fig:ParetoTradeofFasterYOLO}
\end{figure}

\begin{figure*}[h!]
  \centering
   \includegraphics[width=.82\textwidth]{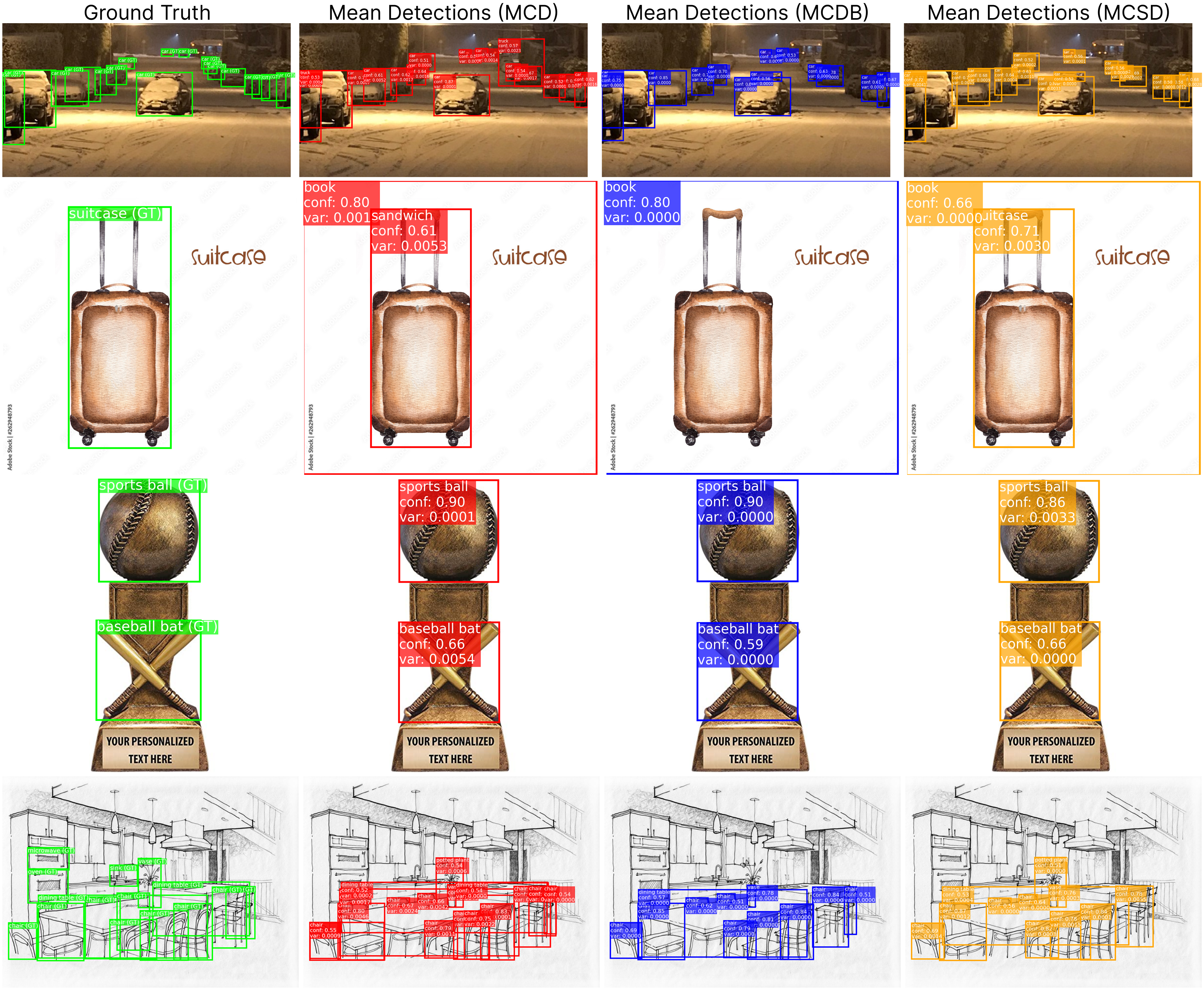}
   \caption{Qualitative comparison of detection and uncertainty for data under distribution shift, using RT-DETRx. Examples from the COCO-O dataset, comparing detection outputs across different uncertainty quantification methods. Each row represents a distinct domain shift: Row 1 (\textit{Weather}), Row 2 (\textit{Painting}), Row 3 (\textit{Handmake}), and Row 4 (\textit{Sketch}). Columns from left to right show: Ground Truth (green bounding boxes), Mean Detections (MCD) (red), Mean Detections (MCDB) (blue), and Mean Detections (MCSD) (orange).}
   \label{fig:OODqualitative}
\end{figure*}

\subsection{Extended Pareto Analysis}
\label{subsec:ext_pareto}

Complementing \cref{fig:ParetoTradeof} in the main text, we provide the complete Pareto trade-off plots for the Faster R-CNN (\cref{fig:ParetoTradeofFasterRCNN}) and YOLOv8x (\cref{fig:ParetoTradeofFasterYOLO}) architectures. 

Consistent with the RT-DETR results discussed in the main paper, we observe that the \textbf{confidence threshold} acts as the dominant variable governing the trade-off between predictive performance (mAP) and uncertainty ranking (AUARC). Higher thresholds generally improve uncertainty ranking at the cost of precision, while lower thresholds maximize mAP but degrade calibration and ranking capabilities.

\vspace{1mm}
\noindent \textbf{Sensitivity to Regularization Magnitude.} In the case of Faster R-CNN (\cref{fig:ParetoTradeofFasterRCNN}), we observe a distinct drop in predictive performance based on the regularization magnitude (drop rate).
We evaluated a wider range of regularization intensities to capture this behavior. The analysis reveals that lower drop rates ($\approx 0.1$) yield the highest mAP, whereas increasing the rate ($\approx 0.3$) significantly degrades predictive performance without a commensurate gain in uncertainty quality.

\vspace{1mm}
\noindent \textbf{Selection of Pareto-Optimal Configurations} To facilitate the tabular comparison in \cref{tab:id_results} of the main paper, we required a principled criterion to select a single representative configuration from the Pareto frontiers shown in \cref{fig:ParetoTradeofFasterRCNN}, \cref{fig:ParetoTradeofFasterYOLO}, and \cref{fig:ParetoTradeof}.

We define the Ideal Performance Point (IPP) as the coordinate combining perfect accuracy and perfect uncertainty ranking, assuming normalized metrics where $\text{mAP} \in [0, 1]$ (higher is better) and $\text{AUARC} \in [0, 1]$ (higher is better).

For each method (MCD, MCDB, MCSD), we selected the configuration that minimizes the Euclidean distance $d$ to this IPP:
\begin{equation*}
    d(c) = \sqrt{(x_{IPP} - \text{AUARC}_c)^2 + (y_{IPP} - \text{mAP}_c)^2}
\end{equation*}
where $c$ represents a specific hyperparameter configuration. This approach balances the two competing objectives, ensuring the selected configuration represents the best overall compromise between predictive power and safety-critical reliability.

\subsection{Qualitative Evaluation}

To complement the quantitative aggregate metrics presented in the main paper, we provide illustrative qualitative examples in \cref{fig:OODqualitative}. These samples are drawn from the COCO-O dataset, representing distribution-shifted scenarios where reliable uncertainty quantification is most critical.

\vspace{1.5mm}
\noindent \textbf{Comparison with Prior Work.} Previous work by Yelleni \etal~\cite{yelleni_monte_2024} posited, based on a limited set of examples, that MCDB is superior to MCD because it eliminates specific misclassifications (e.g., misclassifying a loaf of bread as a sheep). While our visual analysis corroborates that MCDB produces fewer false positive detections (hallucinations or misclassifications) than MCD, we observe that this behavior stems from a significantly more conservative detection threshold.

\vspace{1.5mm}
\noindent \textbf{The Precision-Recall Trade-off.} As illustrated in \cref{fig:OODqualitative}, this conservatism often extends to false negatives. In the first row of the figure (Car scenario), while MCD and MCSD successfully identify the majority of the vehicles despite the domain shift, MCDB suppresses some predictions entirely.
This suggests that the "safety" observed in \cite{yelleni_monte_2024} comes at the cost of reduced recall. In contrast, MCSD maintains a detection density comparable to MCD, thus capturing the true positives while offering the calibration benefits detailed in our quantitative tables. This visual evidence aligns with the quantitative findings where MCDB generally yielded lower mAP scores compared to MCD and MCSD across the tested architectures.

\newpage

\subsection{Sensitivity to Stochastic Layer Placement}

We observed that the efficacy of stochastic regularization for variational inference is dependent on the architectural depth at which it is applied. To quantify this architectural inductive bias, we compare performance when stochastic layers are restricted to the earlier stages of the adjusted blocks against the later stages. \Cref{tab:earlyLate} presents the mean calibration (ECE) and uncertainty ranking (AUARC) metrics across these configurations, verifying the trends reported in the main paper.

\vspace{1.5mm}
\noindent \textbf{Data Inclusion Criteria.} Unlike the Pareto-optimal results reported in the main text (which isolate the single best trade-off configuration), \Cref{tab:earlyLate} reports the mean performance across all configurations that met a minimum viability threshold in mAP and AUARC. Entries marked as \textbf{N/A} indicate configurations that resulted in severe performance degradation, rendering them non-viable for deployment.

It is important to note that because this table averages over a broader set of "viable" models rather than selecting the strict Pareto-best, certain uncertainty metrics (specifically AUARC) may appear higher here than in \cref{tab:id_results} in the main paper.

\begin{table}[t]
\centering
\scriptsize
\setlength{\tabcolsep}{4pt}
\caption{Impact of stochastic layer placement. Comparison of stochasticity applied to earlier vs. later stages in single and multiple (\textit{Half Block}) layers. \textit{N/A} denotes configurations that failed to meet minimum viability thresholds.}
\label{tab:earlyLate}
\begin{tabular}{lllcccc}
\toprule
 & & & \multicolumn{4}{c}{\textbf{Adapted Layers}} \\
\cmidrule(lr){4-7}
& & & \multicolumn{2}{c}{\textbf{Single Layer}} & \multicolumn{2}{c}{\textbf{Half Block}} \\
\cmidrule(lr){4-5} \cmidrule(lr){6-7}
\textbf{Model} & \textbf{Method} & \textbf{Metric} & \textbf{First} & \textbf{Last} & \textbf{First} & \textbf{Last} \\
\midrule
Faster R-CNN & MCD & {\textbf{ECE} $\downarrow$} & \textbf{0.236} & 0.254 & \textbf{0.223} & 0.234 \\
 & & {\textbf{AUARC} $\uparrow$} & \textbf{0.680} & 0.672 & \textbf{0.683} & 0.675 \\
\cmidrule(lr){2-7}
 & MCDB & {\textbf{ECE} $\downarrow$} & \textbf{0.260} & 0.263 & 0.257 & \textbf{0.224} \\
 & & {\textbf{AUARC} $\uparrow$} & 0.572 & \textbf{0.668} & 0.612 & \textbf{0.661} \\
\cmidrule(lr){2-7}
 & MCSD & {\textbf{ECE} $\downarrow$} & \textbf{0.231} & 0.244 & 0.226 & \textbf{0.218} \\
 & & {\textbf{AUARC} $\uparrow$} & \textbf{0.668} & 0.664 & \textbf{0.666} & 0.661 \\
\midrule
YOLOv8x & MCD & {\textbf{ECE} $\downarrow$} & \textbf{0.057} & 0.064 & \textbf{0.056} & 0.059 \\
 & & {\textbf{AUARC} $\uparrow$} & 0.725 & \textbf{0.824} & 0.757 & \textbf{0.781} \\
\cmidrule(lr){2-7}
 & MCDB & {\textbf{ECE} $\downarrow$} & N/A & \textbf{0.060} & N/A & N/A \\
 & & {\textbf{AUARC} $\uparrow$} & N/A & \textbf{0.801} & N/A & N/A \\
\cmidrule(lr){2-7}
 & MCSD & {\textbf{ECE} $\downarrow$} & \textbf{0.044} & 0.064 & \textbf{0.048} & 0.060 \\
 & & {\textbf{AUARC} $\uparrow$} & 0.7890 & \textbf{0.826} & 0.771 & \textbf{0.823} \\
\midrule
RT-DETRx & MCD & {\textbf{ECE} $\downarrow$} & 0.073 & \textbf{0.042} & 0.109 & \textbf{0.077} \\
 & & {\textbf{AUARC} $\uparrow$} & 0.822 & \textbf{0.894} & 0.731 & \textbf{0.818} \\
\cmidrule(lr){2-7}
 & MCDB & {\textbf{ECE} $\downarrow$} & 0.454 & \textbf{0.037} & N/A & N/A \\
 & & {\textbf{AUARC} $\uparrow$} & 0.197 & \textbf{0.897} & N/A & N/A \\
\cmidrule(lr){2-7}
 & MCSD & {\textbf{ECE} $\downarrow$} & 0.063 & \textbf{0.037} & 0.060 & \textbf{0.050} \\
 & & {\textbf{AUARC} $\uparrow$} & 0.874 & \textbf{0.908} & 0.835 & \textbf{0.865} \\
\bottomrule
\end{tabular}
\end{table}

\subsection{Distribution Shift Results}

To supplement the visual trends presented in \cref{fig:ood_shift} of the main paper, \cref{tab:ood_performance} provides the granular quantitative metrics comparing In-Distribution (ID) (COCO) performance against the six distribution shifts in COCO-O.

\vspace{1.5mm}
\noindent \textbf{Quantification of Degradation.} The tabular data highlights the severity of the domain shift. Across all architectures, we observe a precipitous drop in predictive performance as the domain shifts from photorealistic ID (COCO) to abstract representations. For instance, on the 'tattoo' split, which represents one of the most severe shifts, mAP$_{50-95}$ degrades by approximately $76\%$ for Faster R-CNN and $69\%$ for RT-DETRx compared to their in-distribution baselines.

\vspace{1.5mm}
\noindent \textbf{Architecture-Specific Uncertainty Scaling.} A critical insight revealed by \cref{tab:ood_performance} is the disparity in the \textit{magnitude} of the uncertainty response between studied detectors. 
Faster R-CNN exhibits a dynamic range of entropy, nearly doubling from 0.796 (ID) to 1.569 (Tattoo), indicating a robust ability to express epistemic uncertainty. 
In contrast, the dense detectors (YOLOv8x, RT-DETRx) exhibit only a marginal entropy range even when detection performance collapses. While their relative trend is correct (entropy rises as mAP falls), their apparent overconfidence suggests that dense detectors require stronger calibration scaling to be interpretable in safety-critical OOD contexts.

\vspace{1.5mm}
\noindent \textbf{Methodological Robustness.} Finally, the data confirms that MCSD maintains strict parity with MCD under shift. On the hardest split (Tattoo), the performance gap between MCD and MCSD is negligible ($< 0.5\%$ mAP difference on RT-DETRx), confirming that the computational efficiency of MCSD does not come at the cost of OOD robustness.

\subsection{Further Limitations}\label{SupplMat:Limitations}

While MCSD offers a theoretically grounded and empirically robust method for uncertainty quantification, we acknowledge specific limitations inherent to the approach.

First, unlike MCD, which can be applied to almost any neural architecture, MCSD imposes a strict architectural dependency: it requires the presence of residual skip-connections to function as a valid Bayesian approximation.

Second, consistent with all ensemble-based and Monte Carlo methods, MCSD incurs a computational overhead at inference time. Generating a calibrated uncertainty estimate requires T forward passes, scaling the inference latency linearly. While this is acceptable for safety-critical offline analysis, it poses challenges for real-time constraints compared to single-pass deterministic methods.

Finally, our empirical analysis is currently restricted to the COCO and COCO-O datasets. Applying MCSD across a wider variety of object detection datasets is necessary to fully confirm the generalizability of these findings.

\newpage

\makeatletter
\setlength{\@fptop}{0pt}
\makeatother

\clearpage
\onecolumn

\begin{table}[t!]
\centering
\scriptsize
\caption{Quantitative results under distribution shift. Comparison of mAP and Predictive Entropy across the COCO-O domains. Note the difference in Entropy measures between Faster R-CNN (Shannon Entropy) and the dense detectors (Binary Entropy).}
\label{tab:ood_performance}
\begin{tabular}{lllccccccc}
\toprule
 & & & \multicolumn{7}{c}{\textbf{Dataset}} \\
\cmidrule(lr){4-10}
\textbf{Model} & \textbf{Method} & \textbf{Metric} & \textbf{COCO (ID)} & \textbf{weather} & \textbf{painting} & \textbf{handmake} & \textbf{cartoon} & \textbf{sketch} & \textbf{tattoo} \\
\midrule
Faster R-CNN & MCD & {\textbf{mAP.5.95} $\uparrow$} & 0.270 & 0.168 & 0.148 & 0.102 & 0.090 & 0.059 & 0.064 \\
 & & {\textbf{Entropy} $\downarrow$} & 0.796 & 1.101 & 1.289 & 1.505 & 1.374 & 1.310 & 1.569 \\
\cmidrule(lr){2-10}
 & MCDB & {\textbf{mAP.5.95} $\uparrow$} & 0.258 & 0.156 & 0.157 & 0.091 & 0.090 & 0.061 & 0.061 \\
 & & {\textbf{Entropy} $\downarrow$} & 0.894 & 1.155 & 1.340 & 1.531 & 1.375 & 1.371 & 1.546 \\
\cmidrule(lr){2-10}
 & MCSD & {\textbf{mAP.5.95} $\uparrow$} & 0.266 & 0.161 & 0.147 & 0.085 & 0.088 & 0.059 & 0.054 \\
 & & {\textbf{Entropy} $\downarrow$} & 0.646 & 0.944 & 1.100 & 1.26 & 1.134 & 1.111 & 1.375 \\
\midrule
YOLOv8x & MCD & {\textbf{mAP.5.95} $\uparrow$} & 0.499 & 0.413 & 0.382 & 0.261 & 0.211 & 0.201 & 0.156 \\
 & & {\textbf{Entropy} $\downarrow$} & 0.00920 & 0.01005 & 0.01026 & 0.01024 & 0.01040 & 0.01030 & 0.01121 \\
\cmidrule(lr){2-10}
 & MCDB & {\textbf{mAP.5.95} $\uparrow$} & 0.462 & 0.379 & 0.344 & 0.231 & 0.181 & 0.174 & 0.143 \\
 & & {\textbf{Entropy} $\downarrow$} & 0.00958 & 0.01014 & 0.01041 & 0.01055 & 0.01044 & 0.01068 & 0.01105 \\
\cmidrule(lr){2-10}
 & MCSD & {\textbf{mAP.5.95} $\uparrow$} & 0.491 & 0.407 & 0.376 & 0.259 & 0.201 & 0.194 & 0.148 \\
 & & {\textbf{Entropy} $\downarrow$} & 0.00909 & 0.00990 & 0.01017 & 0.01011 & 0.01025 & 0.01026 & 0.01096 \\
\midrule
RT-DETRx & MCD & {\textbf{mAP.5.95} $\uparrow$} & 0.449 & 0.393 & 0.414 & 0.284 & 0.226 & 0.201 & 0.137 \\
 & & {\textbf{Entropy} $\downarrow$} & 0.0936 & 0.0942 & 0.1000 & 0.1141 & 0.1099 & 0.1051 & 0.1247 \\
\cmidrule(lr){2-10}
 & MCDB & {\textbf{mAP.5.95} $\uparrow$} & 0.396 & 0.335 & 0.372 & 0.247 & 0.191 & 0.179 & 0.119 \\
 & & {\textbf{Entropy} $\downarrow$} & 0.0864 & 0.0875 & 0.0931 & 0.1062 & 0.1000 & 0.0973 & 0.1198 \\
\cmidrule(lr){2-10}
 & MCSD & {\textbf{mAP.5.95} $\uparrow$} & 0.437 & 0.380 & 0.395 & 0.263 & 0.217 & 0.200 & 0.135 \\
 & & {\textbf{Entropy} $\downarrow$} & 0.0916 & 0.0939 & 0.0989 & 0.1143 & 0.1071 & 0.1039 & 0.1218 \\
\bottomrule
\end{tabular}
\end{table}